\documentclass[3p,12pt]{elsarticle}

\usepackage{amssymb}
\usepackage{amsmath}
\usepackage{array}
\usepackage{float}
\usepackage{color}
\usepackage{ulem}

\newcolumntype{M}[1]{>{\centering\arraybackslash}m{#1}}
\usepackage{newtxmath}

\journal{Journal of Manufacturing Systems}

\begin{document}

\begin{frontmatter}



\title{Classification of Primitive Manufacturing Tasks from Filtered Event Data}


\author[inst1]{Laura Duarte}
\author[inst1,*]{Pedro Neto}

\affiliation[inst1]{organization={Centre for Mechanical Engineering, Materials and Processes (CEMMPRE)},
            addressline={University of Coimbra},
            postcode={3030-788}, 
            state={Coimbra},
            country={Portugal}}
\affiliation[*]{Corresponding author: pedro.neto@dem.uc.pt}

\begin{abstract}
Collaborative robots are increasingly present in industry to support human activities. However, to make the human-robot collaborative process more effective, there are several challenges to be addressed. Collaborative robotic systems need to be aware of the human activities to (1) anticipate collaborative/assistive actions, (2) learn by demonstration, and (3) activate safety procedures in shared workspace. This study proposes an action classification system to recognize primitive assembly tasks from human motion events data captured by a Dynamic and Active-pixel Vision Sensor (DAVIS). Several filters are compared and combined to remove event data noise. Task patterns are classified from a continuous stream of event data using advanced deep learning and recurrent networks to classify spatial and temporal features. Experiments were conducted on a novel dataset, the dataset of manufacturing tasks (DMT22), featuring 5 classes of representative manufacturing primitives (PickUp, Place, Screw, Hold, Idle) from 5 participants. Results show that the proposed filters remove about 65\% of all events (noise) per recording, conducting to a classification accuracy up to 99,37\% for subjects that trained the system and 97.08\% for new subjects. Data from a left-handed subject were successfully classified using only right-handed training data. These results are object independent.
\end{abstract}




\end{frontmatter}


\section{Introduction}
\label{sec:1}
Collaborative robots play an increasingly important part in the manufacturing landscape. An efficient human-robot collaboration joins together the cognitive coordination, dexterity and flexibility abilities of humans with robot’s accuracy and ease of execution of repeatable tasks \cite{R1,R2}. In a shared workspace, collaborative robotic systems need to be aware of the human activities to (1) anticipate collaborative/assistive actions, (2) learn by demonstration, and (3) activate safety procedures. Robot situational awareness can be implemented at distinct levels, featuring the perception of elements in the current situation, comprehension of the current situation and the prediction of the element’s future states \cite{R3}. This study tackles the first and second levels of the collaborative robot’s situational awareness, i.e., the perception of elements and comprehension of the current situation, in the context of a set of primitive assembly tasks that together compose the complete assembly process of a given object. The recognition and prediction of human actions is key to make the robot aware of human actions in a shared workspace, promoting an effective and safe collaboration \cite{R3-1}. Reference studies rely on object detection and pose estimation from vision-based systems to infer human actions \cite{R3-2, R3-3}, as well as deep learning-based classifiers \cite{R3-4}.

Vision is a sense humans use to perceive the world. As such, machine vision is often favoured in robotic systems to capture information about the environment. Traditionally, machine vision relies on the use of frame-based cameras, which benefit from years of continuous improvements in both equipment and machine vision algorithms. In the last few years, we have witnessed an increase in image resolution, which requires more storage memory and demands complex methods to achieve the classification of human actions. Additionally, frame-based cameras performance suffers from motion-blur, high latencies and low dynamic range. These issues promoted the development of the event camera, an alternative vision sensor inspired by biological vision.

\begin{figure}[]
  \centering\includegraphics[width=0.85\textwidth]{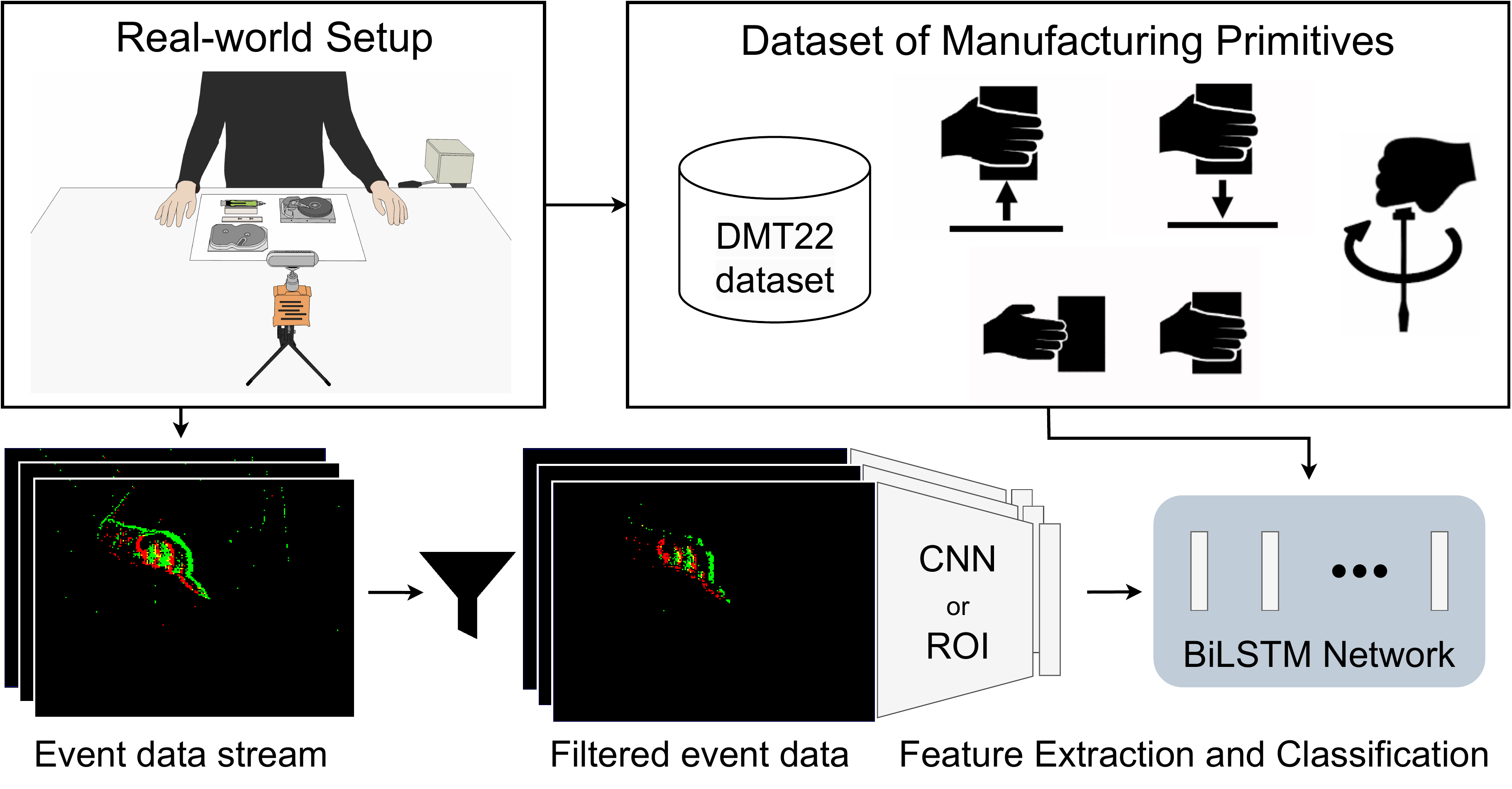}
\caption{The proposed framework to classify primitive manufacturing tasks from filtered event data. Assembly task patterns are classified from a continuous stream of event data through a convolutional neural network (CNN) or region of interest (ROI), after passing several filters to remove event data noise. The classifier, a Long Short-Term Memory (LSTM) network with a single bidirectional layer (BiLSTM) is trained on the proposed dataset of manufacturing tasks (DMT22).}
\label{fig:0}       
\end{figure}

In this study, we propose to capture human actions (primitive assembly tasks) using a Dynamic and Active-pixel Vision Sensor (DAVIS) event camera. It detects changes in brightness in the captured scene, by measuring logarithmic light intensity asynchronously and for each of the sensor’s pixels \cite{R4}, making it suitable to detect motion. As event cameras naturally suffer from random noise, several filters are combined to remove event data noise. The proposed event-based dataset contains five representative assembly primitives. Data were collected from different subjects and assembled objects. Advanced deep learning and recurrent network classifiers combine the classification of spatial and temporal actions. Fig.~\ref{fig:0} shows an overview of the proposed framework. The main contributions of this study are listed as follows:

\begin{itemize}
	\item An efficient method of combining multiple event filters, consistently targeting and removing noise events. The event filters' performance is thoroughly tested on the metrics of data reduction and classification accuracy.
    \item A novel event-based dataset, the dataset of manufacturing tasks (DMT22), featuring five representative assembly primitive actions (PickUp, Place, Screw, Idle and Hold), collected from different subjects and assembled objects. Recorded data are available in an open-source dataset \cite{R29}. No event datasets focused on primitive manufacturing tasks are currently publicly available.
    \item Use of two distinct classification methods to evaluate the event filter's impact on classification accuracy. Different data selection methods further add variety. Comparison between the classification results obtained from a Recurrent Network (RN) architecture that takes as input the region of interest (ROI) features of the human hand captured by the event camera (RN-ROI) and from a deep learning architecture LRCN-TBR, combining a Long-term Recurrent Convolutional Network (LRCN) with the Temporal Binary Representation (TBR) method that converts the camera event output stream into image frames.
    \item Study of the impact of left-handed and right-handed subject data on classification accuracy.
    \item The proposed RN-ROI and LRCN-TBR classifiers are object independent due to the use of primitive actions;
    \item Evaluation of the proposed framework in the assembly of different objects by different subjects working in unstructured environment.
\end{itemize}

\section{Related Studies}
\label{sec:2}
\subsection{Event Camera}
\label{sec:2.1}
\begin{figure}[]
  \centering\includegraphics[width=0.75\textwidth]{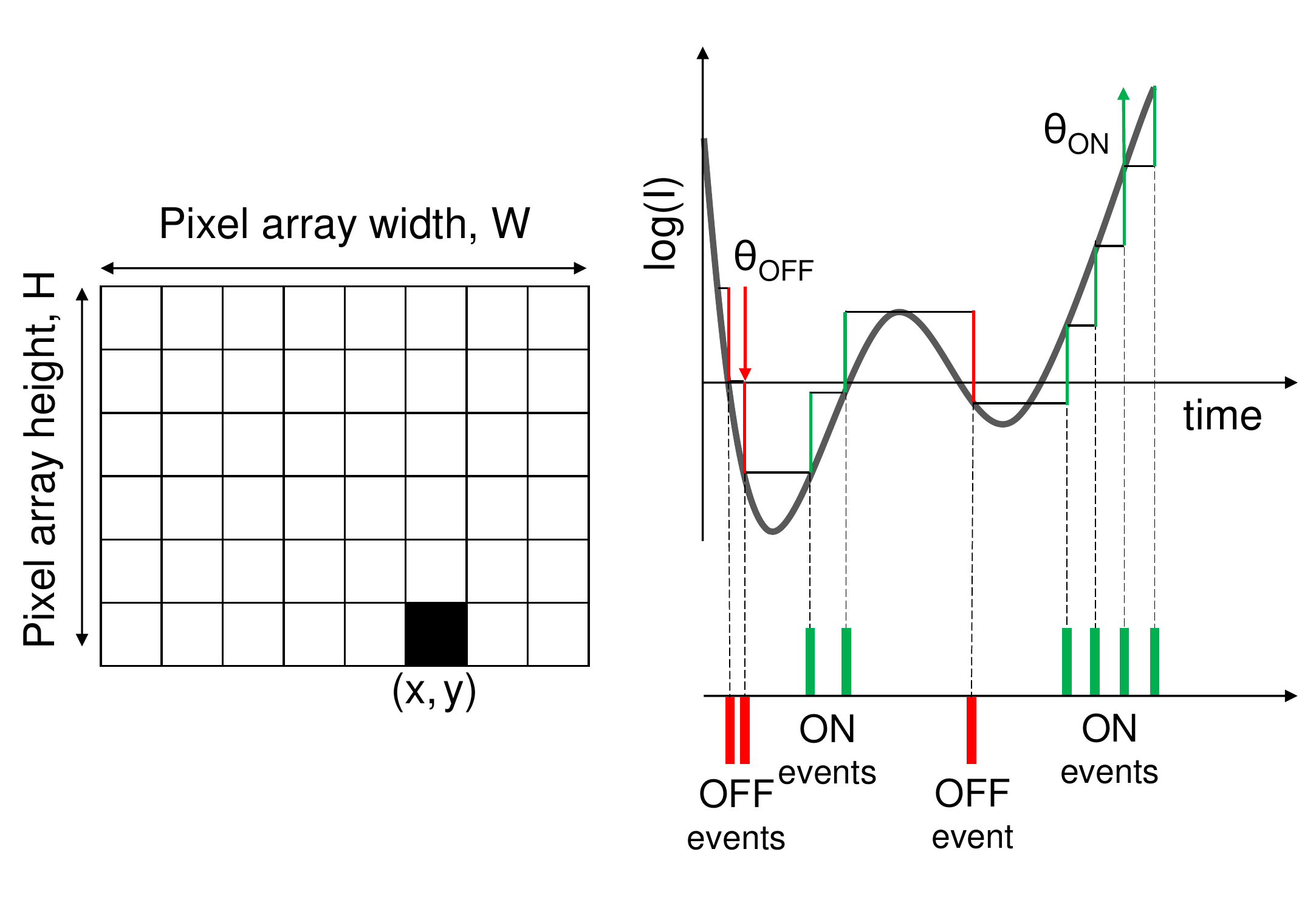}
\caption{Event generated in a DVS pixel. A pixel with coordinates $(x,y)$ belonging to the DVS’s $W\times{H}$ sensor array reads the captured logarithmic light intensity, log(I), over time and responds in the form of ON and OFF events when the upper or lower contrast thresholds, $\theta_{ON}$ and $\theta_{OFF}$ respectively, are exceeded.}
\label{fig:1}       
\end{figure}

One of the most well-known event cameras, the Dynamic Vision Sensor (DVS) \cite{R5}, operates by measuring temporal contrast, which is characterized by changes in brightness in the captured scene. The brightness of a scene is defined by the logarithmic light intensity, log(I). The DAVIS combines the DVS with an active pixel sensor (APS) at the pixel level. In each DVS pixel, the current log(I) is continuously compared to the memorized log(I) value of the same pixel. The pixel generates an asynchronous event when either a lower contrast threshold, $\theta_{OFF}$, or an upper contrast threshold, $\theta_{ON}$, is exceeded, Fig.~\ref{fig:1}. After firing the event, the pixel memorizes the current log(I) value and then resets itself in order to capture the next change in scene brightness. Each generated event, $\boldsymbol{e}_{i}=(x_{i},y_{i},ts_{i},pol_{i})$, is indexed using a unique temporal index $i$ and contains the information about the event’s coordinates in the pixel array, $x$ and $y$, the timestamp of the occurrence of the event, $ts$, and the sign of the associated brightness change, known as the event’s polarity, $pol$. For a positive brightness change an ON event (positive event) is created, $pol=1$, and for a negative brightness change an OFF event (negative event) is created, $pol=0$.

\subsection{Event Filters}
\label{sec:2.2}
\subsubsection{Background Activity Filters}
\label{sec:2.2.1}
Event cameras naturally suffer from random noise, due to thermal noise and leakage currents from the event camera itself \cite{R6}. Background activity noise occurs when a pixel randomly outputs an event without being triggered by a brightness change in the scene.

\begin{figure}[]
  \centering\includegraphics[width=0.650\textwidth]{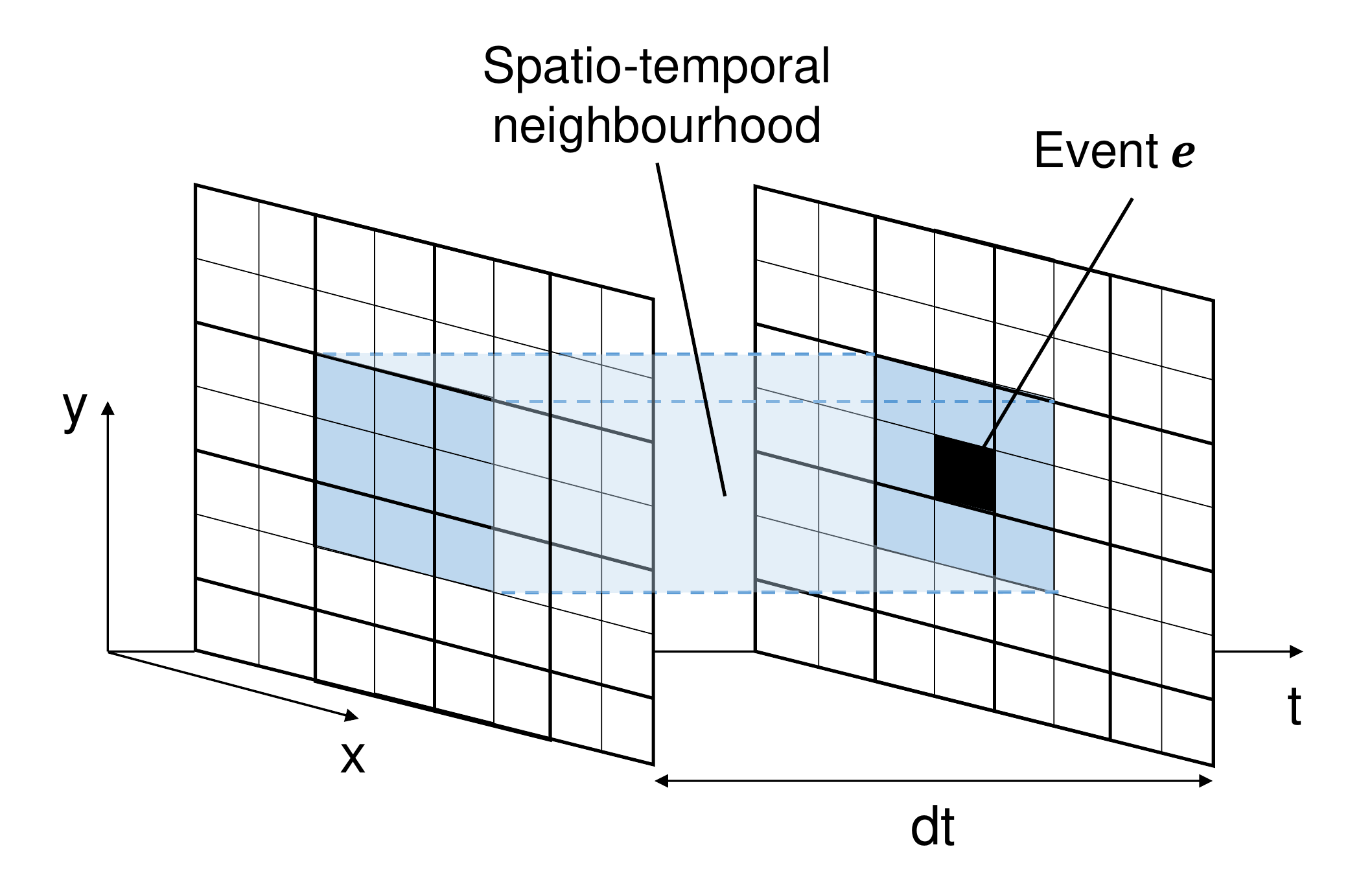}
\caption{Illustration of the spatio-temporal neighbourhood as in \cite{R7}. The event $\boldsymbol{e}$ will only pass through the Background Activity filter if another event exists within its spatio-temporal neighbourhood (blue).}
\label{fig:2}       
\end{figure}

The first attempt at an event-based Background Activity filter registered spatio-temporal neighbours in a timestamp map \cite{R7}. Each new event’s timestamp is stored in all eight neighbouring pixels on the timestamp map, Fig.~\ref{fig:2}. Then, if the new event’s current timestamp exceeds the value of the timestamp map at this event’s location by less than the support time, $dt$, the event goes through the filter. Otherwise, it is discarded. The timestamp map in this method uses one memory cell per sensor pixel, $O(N^2)$\footnote{Space complexity expressed using the $O$ notation, where $N$ represents the camera’s pixel array size. $O(N)=O(W)=O(H)$, with $W$ and $H$ as the DAVIS pixel array width and height, respectively.}.

\begin{figure}[]
  \centering\includegraphics[width=0.7\textwidth]{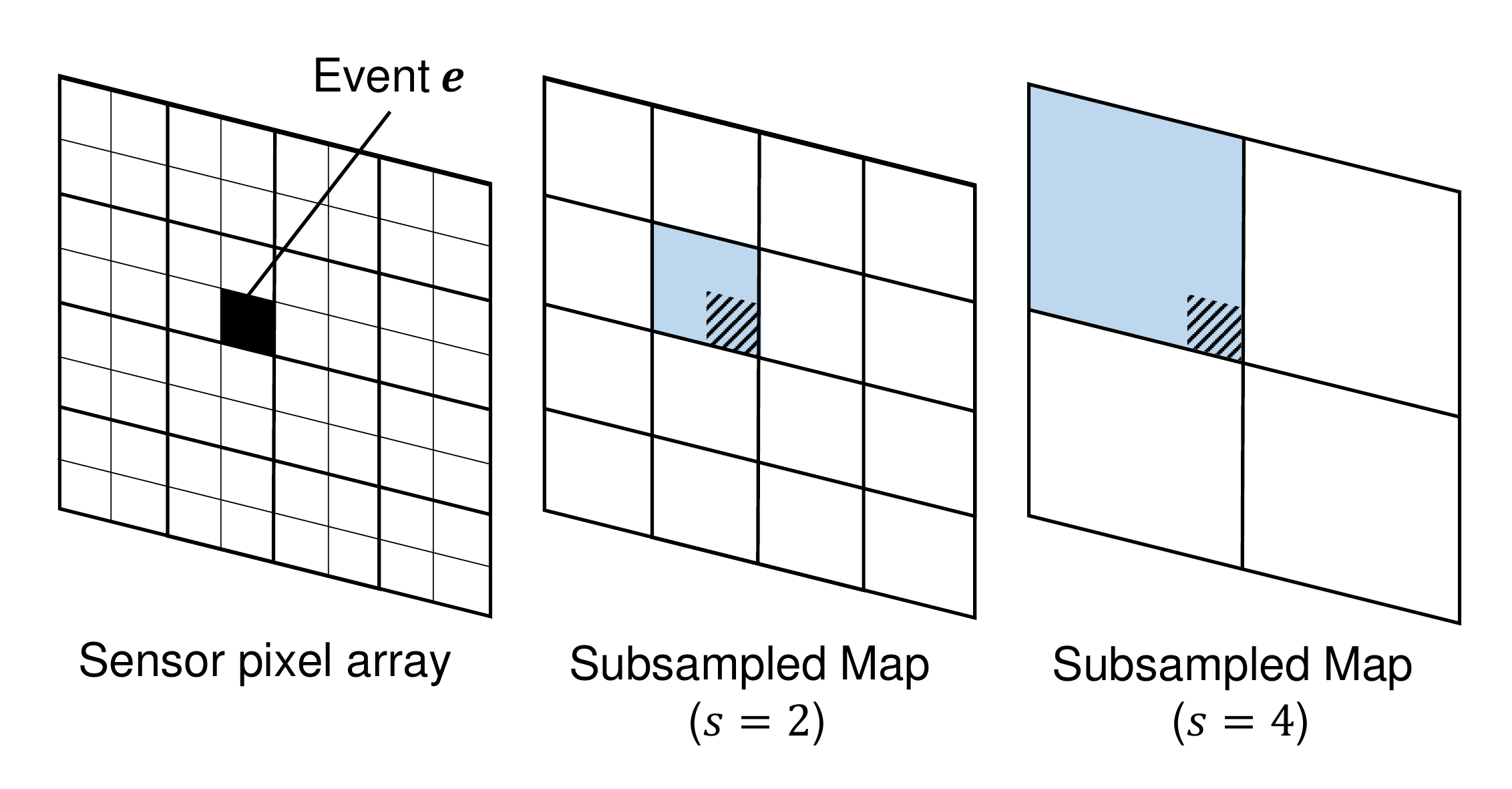}
\caption{Spatio-temporal support (blue) provided by an event on the sensor grid (left) when using a subsampling rate of $s=2$ (middle) or $s=4$ (right).}
\label{fig:3}       
\end{figure}

To reduce the amount of memory required, it was proposed in \cite{R6} to subsample the sensor’s pixel array into groups of $s\times{s}$ pixels, with $s$ as the subsampling rate. This reduces the amount of required memory to $O(N^2/s^2)$ cells. In this case, a new event’s timestamp provides spatio-temporal support to all pixels inside its corresponding group within time $dt$. The main disadvantage of this method is that the spatio-temporal neighbourhood for each event is restricted to the events from the same group. Also, for the method to work properly, small values of $s$ have to be used, such as $s=2$ and $s=4$, as shown in Fig.~\ref{fig:3}.

The Background Activity filter proposed in \cite{R8} challenges both previous attempts by only using a memory complexity of $O(N)$. This is accomplished by using two memory cells per row and two memory cells per column, which store all the necessary data for recovering recent events. By storing information in addition to the timestamp, the error is significantly reduced by allowing correlation with neighbouring events in future time.

Another study uses a hashing-based method \cite{R9}, promoting a further reduction in storage costs. Additionally, this study also presents two metrics for the evaluation of the global performance of Background Activity filters, the percentage of correct predictions for real events and the percentage of real events in the filter’s output. With the DAVIS240 rock-paper-scissors dataset, ROSHAMBO17 \cite{R10}, Liu’s filter \cite{R6} obtains state-of-the-art for these proposed metrics for neighbourhoods of duration $dt\geqslant {1.2} ms$ and a $s=2$ subsampling rate.

\subsubsection{Customized Filters}
\label{sec:2.2.2}
Some event filters must be applied on a case-by-case basis, e.g., to selectively reduce data flow and/or correct some innate issues with the event camera output. A frequent issue with event cameras is that complex circuit bias and manufacturing imperfections cause biased pixels and mismatch contrast threshold among pixels \cite{R11}. An example is the presence of hot pixels, which are pixels with low contrast thresholds that continuously fire events when the visual input is idle. There are two distinct solutions to this issue: (1) record with a stationary event camera and use jAER’s \textit{HotPixelFilter} to identify and then block the pixels with high event rates, and (2) implement an additional rule in one of the previously discussed Background Activity filters which prohibits self-correlation as a pixel’s spatio-temporal support.

Event camera pixels naturally have a refractory period, defined as the duration in time that the pixel ignores brightness changes after an event is generated. The larger the refractory period the fewer events are produced by moving objects \cite{R12}. This type of filter has been used in \cite{R13} to follow the actual position, in time and space, of a helium-filled soap bubble (HFSB) rather than the trace left a few milliseconds after its passage. Alongside fast motion, this filter can also be applied to broad moving edges, along which contrast changes continuously \cite{R14}. In jAER’s \textit{RefractoryFilter} the concept is taken a step further, as it can either filter or pass events within the predefined refractory period. By allowing passage to all events that occur within the refractory period, this filter can work similarly to a self-correlation filter.

The transition between ON and OFF events can be exploited for tracking applications. For example, in \cite{R15}, active LED markers have successfully been used for pose tracking of a quadrotor by analysing these transitions in the DVS data. In \cite{R13}, HFSBs are used as flow tracers, by generating a pixel-sized localized response in the DVS, to reconstruct the 3D path and velocity inside a wind tunnel. To distinguish the events generated by HFSBs and noise, the “pair filter” is used to only pass events which consist of an ON/OFF pair. The jAER \textit{OnOffProximityLineFilter} works similarly, by only outputting events that are supported by a nearby event (neighbour) of the opposite polarity.

The combination of multiple filters is often not fully explored nor detailed in literature. This paper identifies some of the most relevant filter algorithms and builds a novel framework featuring event filters with shared computational resources.

\subsection{Action classification}
\label{sec:2.3}
Gestures are action primitives frequently used as a natural human-robot interface. A set of 25 gestures, used to interface robots, are recognized using two 3D Convolutional Neural Networks (CNNs) and a Long Short-Term Memory (LSTM) network in \cite{R20}. To teach assembly tasks, multiple gestures can be used sequentially and classified as either directional (Up, Down, Left, Right), orientational (90º, 180º), manipulation (Install, Remove, PickUp, Place), and feedback type (Confirm, Stop) \cite{R21}. Human action sequences can be segmented into primitive tasks and then recognized by learning the relation between detected objects and human pose from RGB-D data through the use of graph networks \cite{R22}. Example action classes considered are Idle, Place, Hold, Pour (Kitchen context) and Screw (Workshop context). Human actions can also be divided into a set of static gestures (SGs) and dynamic gestures (DGs) \cite{R23}. An interesting application of segmenting human activity into a sequence of individual actions is presented in \cite{R24}, where a robot can detect forgotten actions and then remind the human of such actions. Robots themselves can also benefit from using primitive tasks to adapt to build complex behaviours. These primitives are usually taught through human demonstration and can then be combined to perform product assembly \cite{R25,R26}. There is a relatively small number of event-based datasets for action and gesture recognition, Table~\ref{tab:1}. None of the datasets available is focused on primitive manufacturing tasks. Primitive actions such as pick up, place, screw, etc. cover the most common manufacturing assembly tasks and can be used sequentially to perform complex assembly processes \cite{R3-1,R3-3,R3-4}. 

\begin{table}[]
\caption{Event-based datasets for action and gesture recognition.\label{tab:1}}
\centering
\begin{tabular}{|M{3cm}|M{1cm}|M{3cm}|M{3cm}|M{4cm}|}
\hline 
{\footnotesize{}Author} & {\footnotesize{}Year} & {\footnotesize{}Dataset Name} & {\footnotesize{}Event Camera Model} & {\footnotesize{}Content Description}\tabularnewline
\hline 
\hline 
{\footnotesize{}A. Amir \cite{R16}} & {\footnotesize{}2017} & {\footnotesize{}DvsGesture} & {\footnotesize{}DVS128} & {\footnotesize{}11 hand gestures under 3 lighting conditions}\tabularnewline
\hline 
{\footnotesize{}I. Lungu \cite{R10}} & {\footnotesize{}2017} & {\footnotesize{}ROSHAMBO17} & {\footnotesize{}DAVIS 240} & {\footnotesize{}3 Rock-Paper-Scissors hand gestures}\tabularnewline
\hline 
{\footnotesize{}S. Baby \cite{R17}} & {\footnotesize{}2018} & {\footnotesize{}DVS Gesture data} & {\footnotesize{}DVS128} & {\footnotesize{}10 hand gestures}\tabularnewline
\hline 
{\footnotesize{}E. Ceolini \cite{R18}} & {\footnotesize{}2020} & {\footnotesize{}DVS-EMG Dataset} & {\footnotesize{}DVS128} & {\footnotesize{}5 sign language gestures. Additional APS frames and EMG data}\tabularnewline
\hline 
{\footnotesize{}S. Innocenti \cite{R19}} & {\footnotesize{}2020} & {\footnotesize{}MICC-Event Gesture Dataset} & {\footnotesize{}Prophesee GEN3S VGA-CD} & {\footnotesize{}Extension of DvsGesture under more challenging conditions}\tabularnewline
\hline 
\end{tabular}
\end{table}

\begin{figure}[]
  \centering\includegraphics[width=0.95\textwidth]{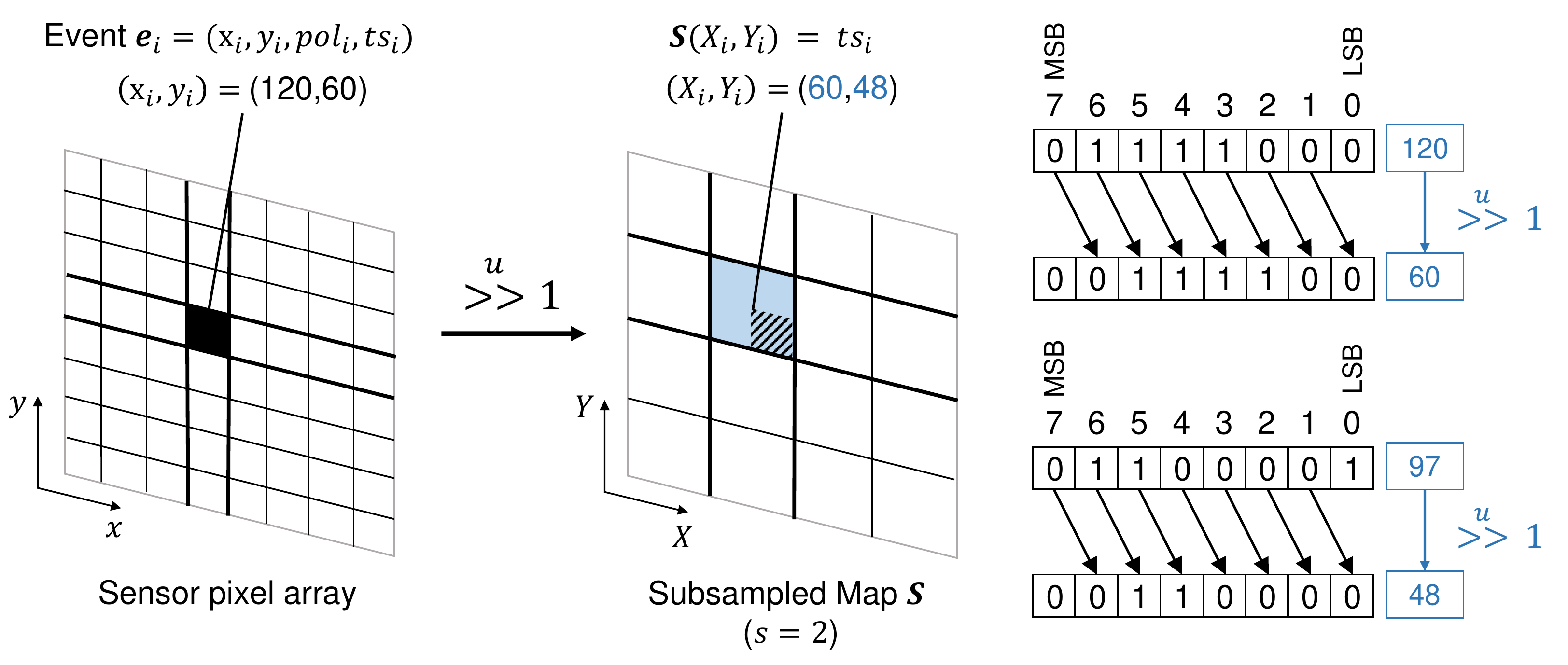}
\caption{A single right unsigned bit shift, $\overset{u}{>>}1$, is used to convert event’s coordinates into subsampling map coordinates. MSB and LSB stand for the most significant bit and least significant bit, respectively.}
\label{fig:4}       
\end{figure}

\section{Methodology}
\label{sec:3}
\subsection{Event Filters}
\label{sec:3.1}
As previously mentioned, the use of subsampling as a basis for a Background Activity filter achieves good results. By subsampling the sensor’s pixel array into groups of $s\times{s}$ pixels, a subsampled map is created, in which the most recent timestamp (from the last event in the corresponding subsampled group) is stored. 
A right unsigned ($u$) bit shift, $\overset{u}{>>}n$, is a bitwise operation which shifts the operand by $n$ bits to the right. In this operation, excess bits shifted off to the right are discarded and the vacant bit positions from the left are filled with zeros. Unsigned bit shifts are a efficient way to perform the division of unsigned integers by powers of two.
A single right logical bit shift, $\overset{u}{>>}1$, can be used to directly map events from the sensor pixel array to the subsampled map corresponding to $s=2$, Fig.~\ref{fig:4}. The only drawback of this operation is that the subsampling rate $s$ must be a power of two, ${s: s = 2^n,n\in\mathbb{N}}$. Assuming a new event, defined as $\boldsymbol{e}_{i}=(x_{i},y_{i},ts_{i},pol_{i})$, the coordinates $x_{i}$ and $y_{i}$ are shifted by $n$ bits into the corresponding subsampled map coordinates, $X_{i}$ and $Y_{i}$:

\begin{equation}
\begin{split}
{X_i}=x_{i}\overset{u}{>>}n \\ 
{Y_i}=y_{i}\overset{u}{>>}n
\end{split}
\end{equation}

\noindent After each new event $\boldsymbol{e}_{i}$ passes through all the filters, or is caught by any one of the filters, its timestamp $ts_{i}$ is stored into the subsampled map $\boldsymbol{S}$:

\begin{equation}
\boldsymbol{S}(X_{i},Y_{i})=ts_{i}
\end{equation}

\noindent To avoid spatio-temporal correlation with non-events, the following condition, called the First Event filter, prohibits correlation with the zero-initialized subsampled map $\boldsymbol{S}$, by only passing events if: 
\begin{equation}
\boldsymbol{S}(X_{i},Y_{i})>0
\end{equation}

\noindent Since most initial events from a recording are noise, the First Event filter substantially reduces the initial spike of noise events despite its finite filtering capacity. In the Background Activity filter, the value of the subsampling map $\boldsymbol{S}(X_{i},Y_{i})$ is compared to the current event’s timestamp, $ts_{i}$, to evaluate if their difference is less than the pre-defined support time $dt_{BA}$. The new event $\boldsymbol{e}_{i}$ is filtered if: 

\begin{equation}
ts_{i}-\boldsymbol{S}(X_{i},Y_{i})\geqslant{dt}_{BA}
\end{equation}

When using a Background Activity filter, a convenient process to filter hot pixels is to negate self-correlation. This means that if the only spatio-temporal neighbour for a new event is an older event at the same pixel coordinates, the new event is deemed invalid. However, spatial information of the events is lost when subsampling occurs, because only the timestamp is stored at the subsampling map $\boldsymbol{S}$. As such, to build a Hot Pixel filter, a new map called the coordinate map $\boldsymbol{C}$, with the same dimension as $\boldsymbol{S}$, will store the coordinates of the last event, $x_{i}$ and $y_{i}$, at $\boldsymbol{C}(X_{i},Y_{i},0)$ and $\boldsymbol{C}(X_{i},Y_{i},1)$, respectively. Thus, an event will only pass the Hot Pixel filter if:

\begin{equation}
\boldsymbol{C}(X_{i},Y_{i},0)\neq{x}_{i}\land{\boldsymbol{C}}(X_{i},Y_{i},1)\neq{y}_{i}
\end{equation}

\noindent After each new event $\boldsymbol{e}_{i}$ is either caught by or passed through any of the event filters, its $x$ and $y$ coordinates are stored in the coordinate map $\boldsymbol{C}$:

\begin{equation}
\begin{split}
\boldsymbol{C}(X_{i},Y_{i},0)=x_{i} \\ 
\boldsymbol{C}(X_{i},Y_{i},1)=y_{i}
\end{split}
\end{equation}

A similar approach to the Background Activity filter is used to build the Refractory filter, but instead, it uses a smaller support time, $dt_{Refr}$ ($dt_{Refr}<dt_{BA}$), and filtering $\boldsymbol{e}_{i}$ if:

\begin{equation}
ts_{i}-\boldsymbol{S}(X_{i},Y_{i})\leqslant{dt}_{Refr}
\end{equation}

\noindent If this condition is met, this means that an event is most likely noise related to flickering. Assuming that both general background activity and hot pixels have been filtered out, the only non-essential information left on the image is the shadow cast (hand and object in this study). These events are not previously filtered due to the fact they create spatio-temporal neighbourhoods which pass through the Background Activity filter. These events have, generally, the same polarity.
By checking neighbouring subsample groups for spatio-temporal support from events of the opposite polarity, it is possible to determine if events are part of a shadow. Taking inspiration from the Background Activity filter method, the Polarity filter will use a similar search method, but in a wider spatio-temporal neighbourhood. A range of coordinates is used instead of a single comparison, and a greater support time, $dt_{Pol}$ ($dt_{Pol}> dt_{BA}$), is used. However, only neighbouring events with opposite polarity, 1 – $pol_{i}$, are used to validate the condition. The event passes the filter if, for any possible coordinate combination within range, the following condition is true:

\begin{equation}
\begin{split}
ts_{i}-\boldsymbol{S}(range_{x},range_{y},1-pol_{i})\leqslant{dt}_{Pol}, \\
\forall{range}_{x}\in{\{{X}_{i}-1,{X}_{i},{X}_{i}+1\}},\forall{range}_{y}\in{\{{Y}_{i}-1,{Y}_{i},{Y}_{i}+1\}}
\end{split}
\end{equation}

\noindent where $\boldsymbol{S}(X_{i},Y_{i},1)$ and $\boldsymbol{S}(X_{i},Y_{i},0)$ are the subsampled maps for positive and negative events, respectively. However, the reduced filtering capacity this filter offers is outweighed by its main issue of requiring the previous filter’s conditions to check for both polarities. The polarity filter mostly serves the purpose of filtering shadows. However, since shadows are not prominently featured in the proposed dataset, this filter will not be used in this study.

\begin{figure}[]
  \centering\includegraphics[width=0.9\textwidth]{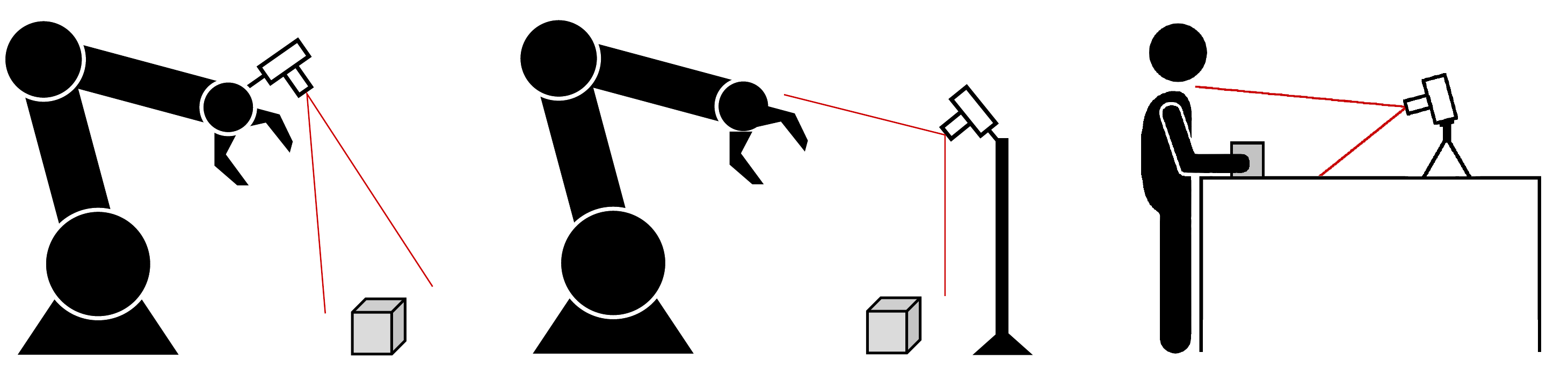}
\caption{Eye-in-Hand system configuration (left), Eye-to-Hand system configuration (middle), and the proposed camera setup (right).}
\label{fig:5}       
\end{figure}

\subsection{Dataset of Manufacturing Tasks (DMT22)}
\label{sec:3.2}
The DMT22 dataset contains recorded data from an event camera, a depth camera and a magnetic tracking system. Recorded data are available in an open-source dataset \cite{R29}. While the magnetic tracker is attached to the human body, the cameras are stationary. In this study, we focus on data from the event camera. The dataset includes data from other sensors so that it can be used by a broader audience and results from different sensor data can be compared.

\begin{figure}[]
  \centering\includegraphics[width=0.75\textwidth]{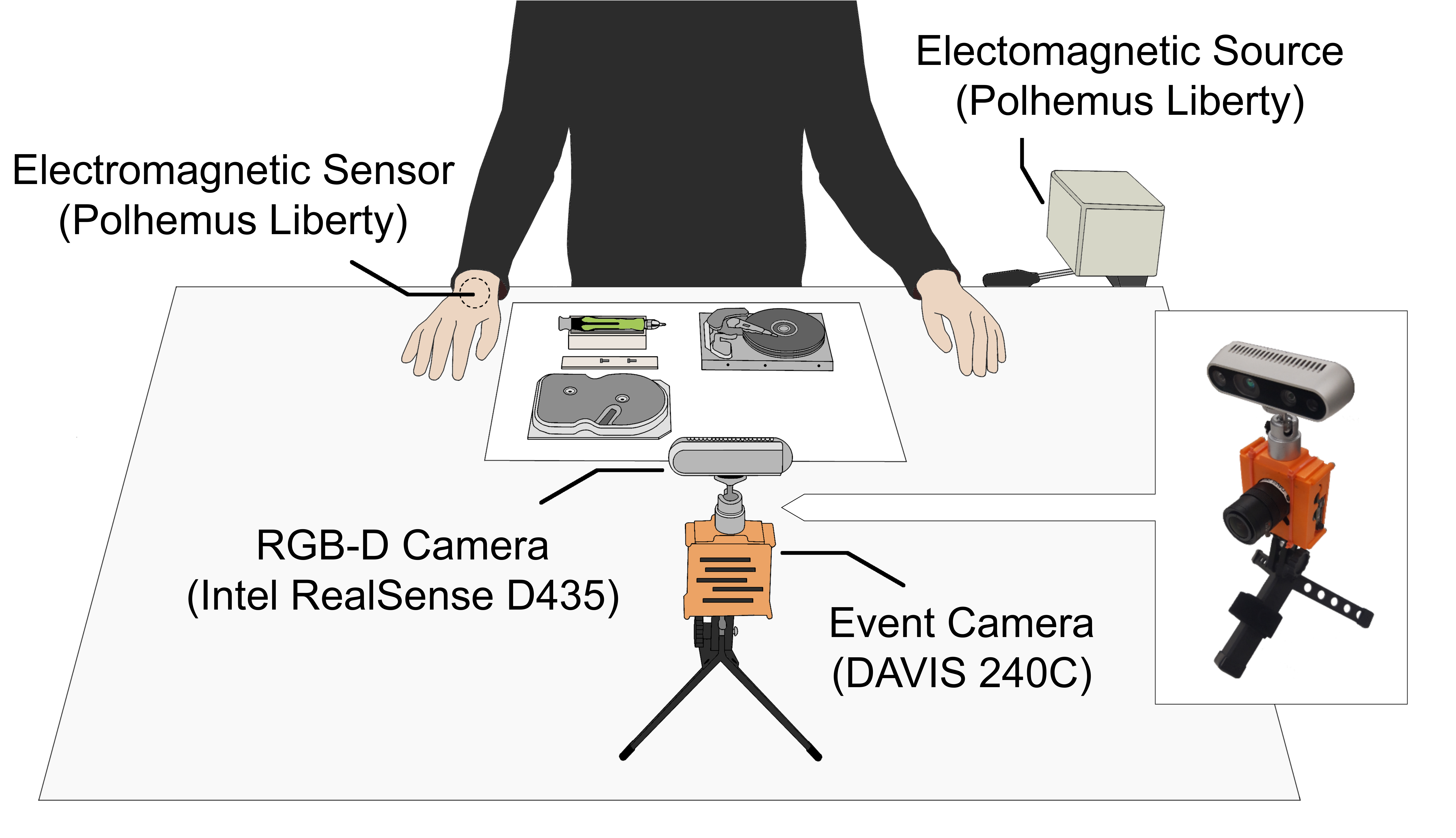}
\caption{The DMT22 dataset acquisition setup featuring the hard disk drive assembly task. The DAVIS event camera and the RGB-D camera are coupled to capture scene data. The electromagnetic sensor is attached to the subject's wrist.}
\label{fig:6}       
\end{figure}

Focusing on the event camera, if it is moving, information about both moving and stationary elements can be obtained, due to the relative movement between them. This is the Eye-in-Hand system, where a camera can be held by a robot with pre-programmed movements to track the camera’s exact pose in 3D space, Fig.~\ref{fig:5} (left). With this, a scene can be viewed from different angles, obtaining information which can be used to reconstruct it in the world coordinate system, as witnessed frequently in Simultaneous Localization and Mapping (SLAM) applications. An alternative is to use an Eye-to-Hand system, in which the camera is placed at a fixed point in the workspace, Fig.~\ref{fig:5} (middle). It facilitates the computation of transformations between image and world coordinate systems through calibration. When the camera is fixed, stationary elements will not be registered in events data, only the elements (human hand) moving relative to the camera will be captured. Thus, the Eye-to-Hand configuration is the more suitable configuration for capturing the manufacturing primitive actions in this dataset, Fig.~\ref{fig:5} (right). The setup used for the acquisition of the dataset data is in Fig.~\ref{fig:6}.

\begin{figure}[]
  \centering\includegraphics[width=1\textwidth]{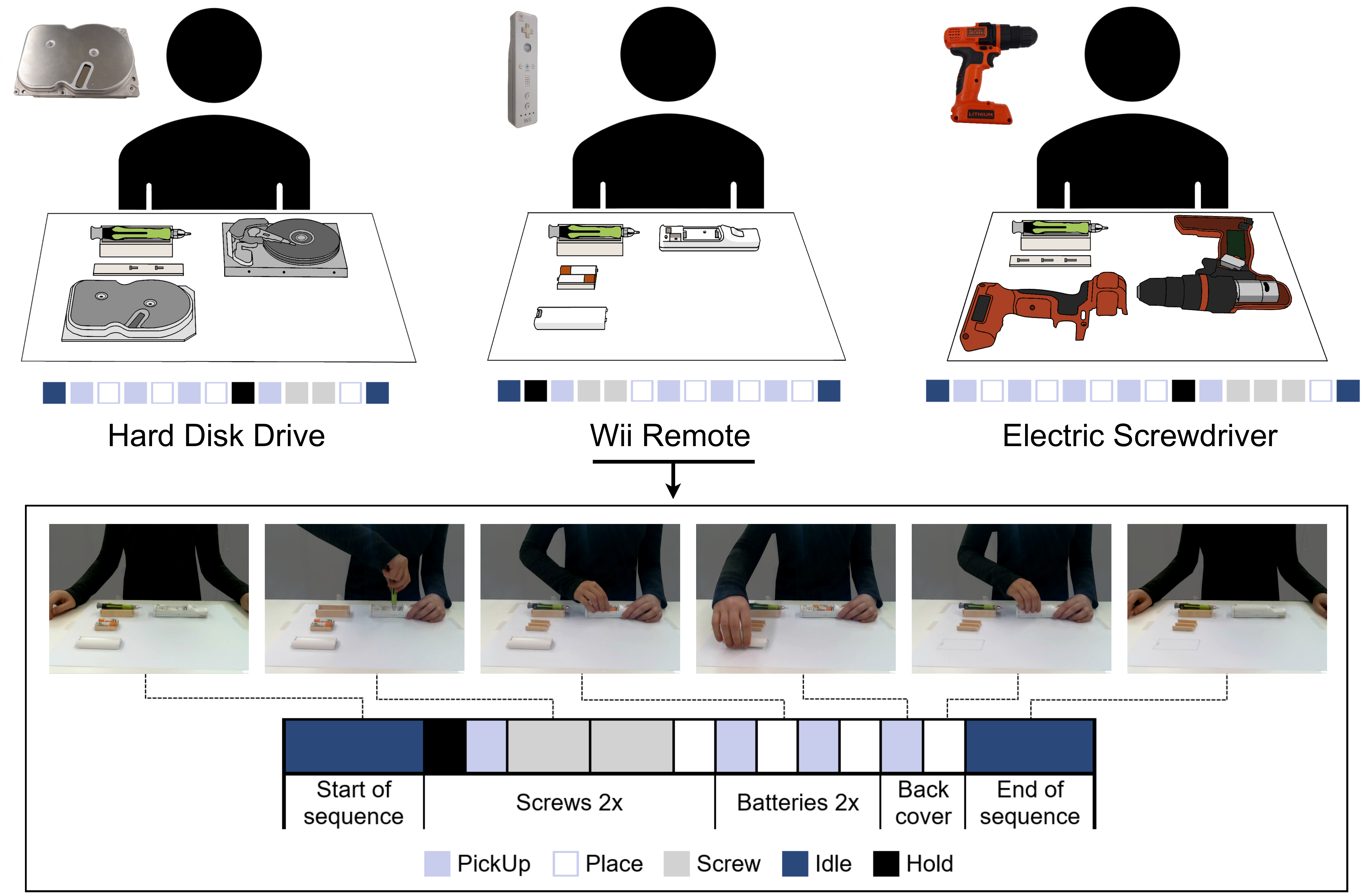}
\caption{The three different objects on the DMT22 dataset and the corresponding sequence of primitive tasks to complete the assembly of each one. Segmentation of a Wii remote assembly task from the DMT22 dataset, with each colour depicting a unique primitive action. It is composed by 13 different primitive tasks: (1) the human is idle, (2) holds the screwdriver, (3) picks up the screwdriver, (4) performs a screw action, (5) performs another screw action, (6) places the screwdriver, (7) picks up a battery, (8) places the battery, (9) picks up another battery, (10) places the battery, (11) picks up the back cover, (12) places the back cover, and (13) ends the assembly idle.}
\label{fig:7}       
\end{figure}

The dataset includes a set of manufacturing assembly tasks, each featuring a different object, a Wii remote, a hard disk drive, and an electric screwdriver, Fig.~\ref{fig:7}. \textcolor{blue}{The objects were selected to be of equivalent complexity. All possess similar primitive tasks, including both “screwing” and “fitting” actions, which are key primitives in assembly tasks. Electro-mechanical items were chosen as they provide a balance between being a complex assembly case and being big enough that its components will be visible on the recordings.}

The tasks were recorded by five subjects: 1 right-handed female subject (Subject 1) and 4 male subjects, of which 3 are right-handed (Subjects 2, 3 and 4) and 1 is left-handed (Subject 5). The selected subjects vary in gender, assembly experience and feature different dominant hands. All subjects were given a brief introduction to each assembly task before recording. Each subject performed all three assembly tasks four times. An exception was created for the left-handed subject, Subject 5, which performed every task both with a right-sided (right-handed) and left-sided (left-handed) placement of the parts to be assembled and the required tools. The right-sided setup was used for all right-handed subjects. The left-sided setup is a mirrored version of the right-sided setup, which means most parts and tools are on the left side of the subject, a more convenient placement for a left-handed subject. These two setups allow to study how well a left-handed subject action can be classified in a context where the classifiers are trained with right-handed subject’s data. The dataset features a total of 72 recordings: 4 right-handed subjects $\times$ 3 tasks $\times$ 4 performances + 1 left-handed subject $\times$ 2 configurations $\times$ 3 tasks $\times$ 4 performances. Each task recording features multiple primitive tasks and lasts approximately 25 seconds.

The data were manually labelled by a single annotator, as to minimize attribution discrepancies. Action classes were labelled as one of the following five primitives: PickUp, Place, Screw, Idle and Hold. Fig.~\ref{fig:7} shows a detailed description of the assembly of the Wii remote consisting of 13 different primitive tasks.
The event camera used to record the dataset is a DAVIS 240C, capturing events with a resolution of 240 pixels $\times$ 180 pixels. DAVIS APS data are captured at 30 fps (frames per second), along with RGB-D data from the Intel RealSense Depth Camera D435 at 30 fps. The Polhemus Liberty electromagnetic tracker was used to capture pose data, at 20 Hz, from a sensor attached to the subject’s wrist. Such data are expected to suffer from magnetic distortion. 

\subsection{Classification Methods}
\label{sec:3.3}
 Two classification methods are proposed to classify the primitive manufacturing/assembly tasks from filtered data:
\begin{enumerate}
  \item RN-ROI: A recurrent network architecture that takes as input features from the region of interest (ROI) of the human hand. 
  \item LRCN-TBR: A deep learning architecture combines a Long-term Recurrent Convolutional Network (LRCN) with the Temporal Binary Representation (TBR) method \cite{R19} that converts the camera event output stream into image frames.
\end{enumerate}

\begin{figure}[]
  \centering\includegraphics[width=1\textwidth]{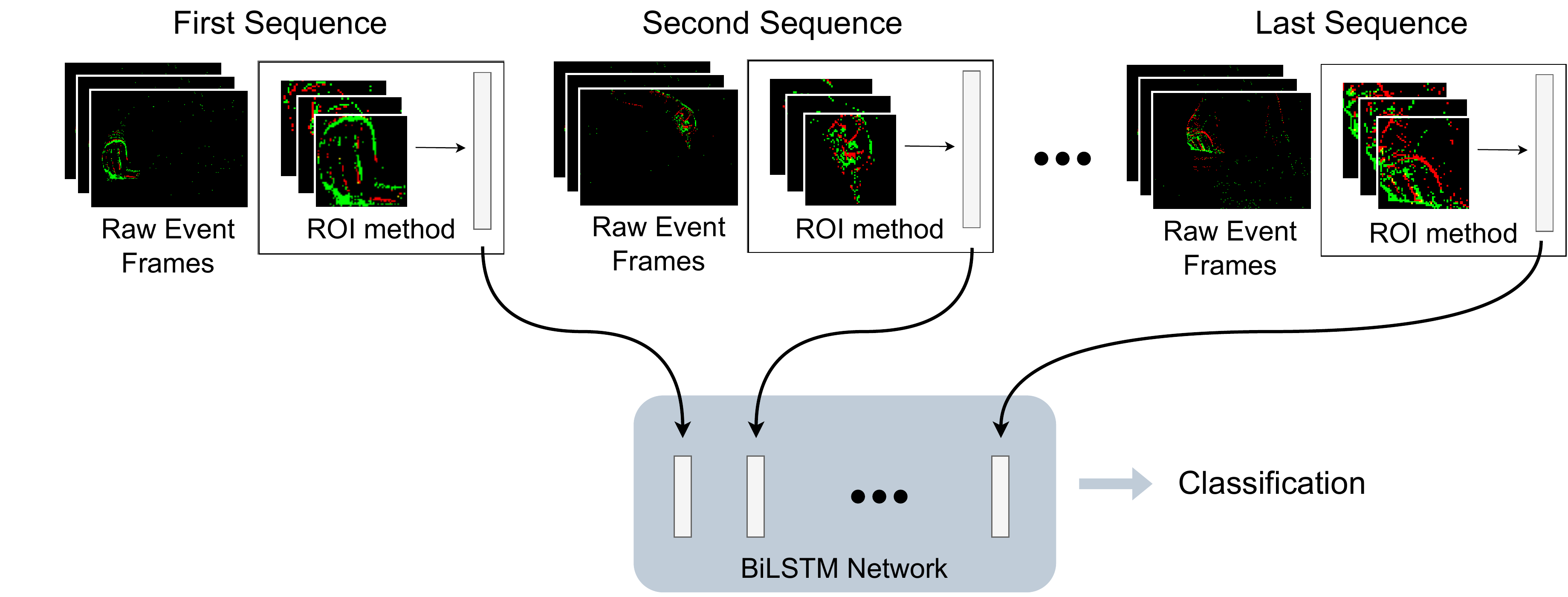}
\caption{Classification architecture of event data using the RN-ROI method.}
\label{fig:9}       
\end{figure}

The RN-ROI method relies on a lightweight algorithm to define the ROI features of the human hand through object edge event activity \cite{R27}. The algorithm detects the pixel columns and rows featuring event activity to define a square ROI containing the human hand. The relevant features from the ROI are its centre coordinates, size, and percentage of active pixels. In \cite{R27}, features are obtained at every 3000 events. In this study, a comparison will be made between the filtered and unfiltered data from the DMT22 dataset. As such, features must be obtained at fixed intervals of time for a fair comparison, so the same “frame” is represented in both analyses. The chosen interval of time to obtain features is $\Delta{t} = 30 ms$. Since the ROI features are presented sequentially and the intent is to use that temporal continuity to classify the primitive actions, a Long Short-Term Memory (LSTM) network with a single bidirectional layer (BiLSTM) is used. Features are zero-centred before classification. The schematic of the network is illustrated in Fig.~\ref{fig:9}. Since the DMT22 dataset is relatively small, a data augmentation technique is used to better train the LSTM network. All data are used in their original state as well as their mirrored state (horizontally flipped), effectively doubling the data available for training the network. This was also done to help the left-sided setup of Subject 5 to better conform to the rest of the DMT22 data.

\begin{figure}[]
  \centering\includegraphics[width=1\textwidth]{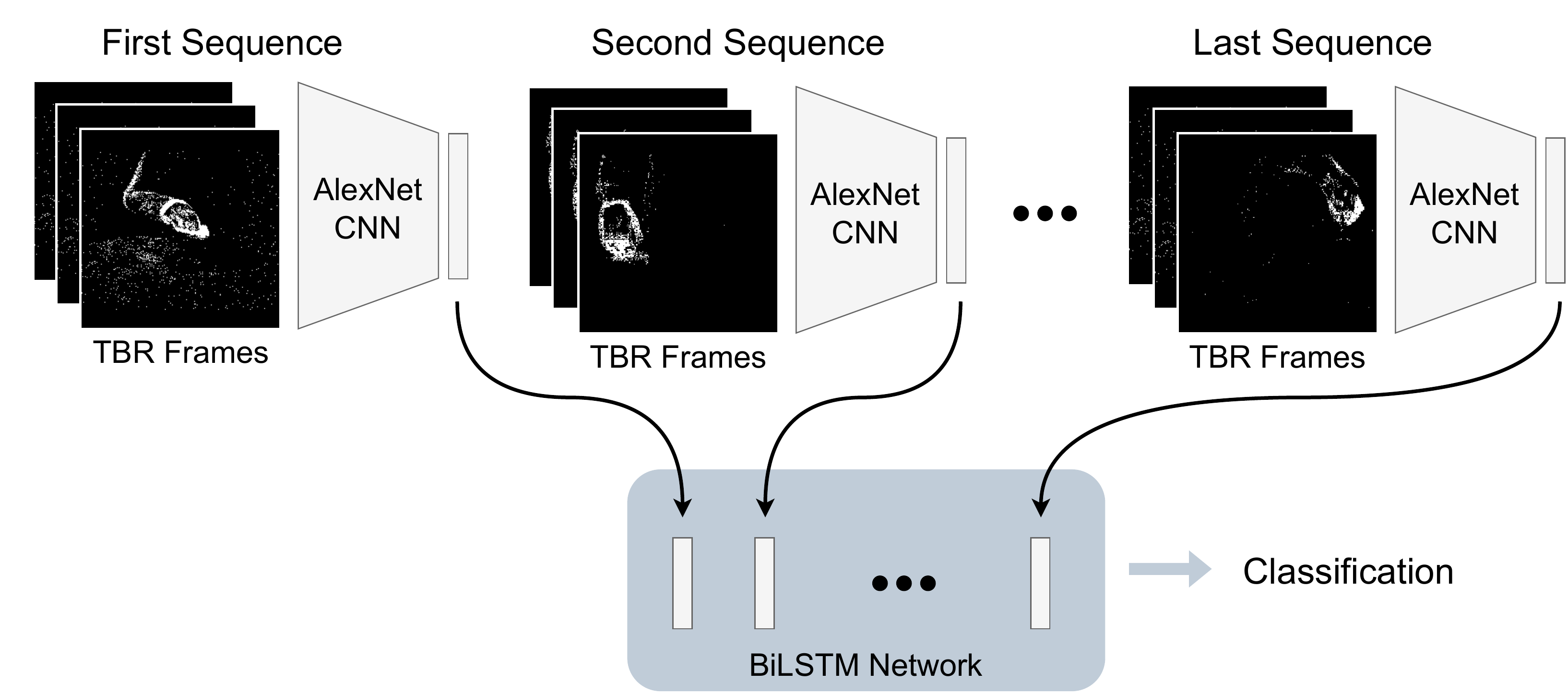}
\caption{Classification architecture of event data using the LRCN-TBR method.}
\label{fig:10}       
\end{figure}

For the LRCN-TBR method, the camera event output stream is converted into image frames. Given an interval of time $\Delta{t}$, a binary representation $b^i$ is built for each pixel. For each possible event coordinate $(x,y)$, the value of the representation is $b_{x,y}^{i}=1$ in case any event is present and $b_{x,y}^{i}=0$ otherwise. By considering a sequence of eight consecutive binary representations, these can be combined into a single 8-bit array, which is converted into a decimal number. The range of possible values corresponds to the greyscale range of $0-255$. As such, this method is an efficient way to create images from events. The images created through this method can not immediately be processed by a conventional LSTM network. Using Convolutional Neural Network (CNN) AlexNet, which is already trained, and by removing its last layers responsible for classification, the CNN is used to obtain features instead of performing classification. The input frames must comply with the expected input size of the chosen CNN, AlexNet. The images created must first be cropped horizontally to a size of $227\times{180}$, to which vertical zero-padding up to $227\times{227}$ is then applied. As a final step, a LSTM network, identical to the one proposed for the RN-ROI method, can then obtain the temporal correlation from the analysed frames to classify the data. The schematic of the LRCN-TBR method is shown in Fig.~\ref{fig:10}.

\section{Experiments and Results}
\label{sec:4}

Detailed information on the experimental setup and data collected is presented in section \ref{sec:3.2}. The assembled objects are described, as well as the subjects, experimental trials, data formats and equipment. The video that accompanies this article shows sample data collected for each assembled object.

\subsection{Filter performance}
\label{sec:4.1}
The filtering results were obtained by applying them to the DMT22 dataset and the Event Data for Hand Tracking EDHT21 dataset \cite{R28}. The filters were built and tested in the open-source Java software framework jAER\footnote{Available at: http://jaerproject.net}.
For each action sequence of the dataset, the event filters performance is measured by comparing the number of total events received to the number of filtered events. 

\begin{equation}
\text{\% of Filtered Events}=\frac{\text{\# of events filtered}}{\text{\# of events in sequence}}
\end{equation}

Filters are applied to event data sequentially, considering the parameters $dt_{BA}=1.5 ms$ and $dt_{Refr}=10 \mu{s}$ for the Background Activity filter and Refractory filter, respectively. Results in Table~\ref{tab:2} were obtained from the DMT22 dataset for two distinct subsampling rates, $s=2$ and $s=4$. Subsampling rates have a high impact on filter performance.

\begin{table}[]
\caption{Average results of event filters on Subject 1 data from the DMT22 dataset.\label{tab:2}}
\centering
\begin{tabular}{|M{3cm}|M{3cm}|M{3cm}|M{4cm}|}
\hline 
{\footnotesize{}Object} & {\footnotesize{}Avg. \% of filtered events for $s=2$} & {\footnotesize{}Avg. \% of filtered events for $s=4$} & {\footnotesize{}Avg. number of events in sequence}\tabularnewline
\hline 
\hline 
{\footnotesize{}Hard Disk Drive} & {\footnotesize{}85.89} & {\footnotesize{}67.39} & {\footnotesize{}661874}\tabularnewline
\hline 
{\footnotesize{}Wii Remote} & {\footnotesize{}82.58} & {\footnotesize{}63.59} & {\footnotesize{}478685}\tabularnewline
\hline 
{\footnotesize{}Electric Screwdriver} & {\footnotesize{}82.27} & {\footnotesize{}62.15} & {\footnotesize{}890886}\tabularnewline
\hline 
\end{tabular}
\end{table}

\begin{figure}[]
  \centering\includegraphics[width=0.8\textwidth]{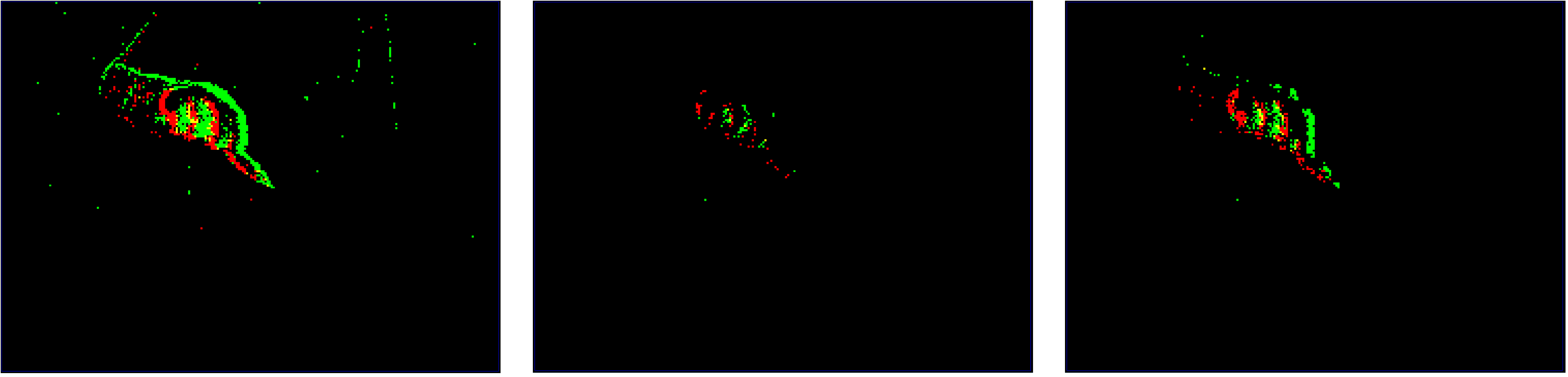}
\caption{Loss of information depending on subsampling rate. Original with no subsampling (left), with subsampling rate of $s=2$ (middle) or with subsampling rate of $s=4$ (right).}
\label{fig:11}       
\end{figure}

Although the results have shown a higher filtering rate with a subsampling rate of $s=2$, relevant features present in the event data are lost, Fig.~\ref{fig:11}. These features are essential for the recognition and classification of data-based action patterns. Filtering with $s=4$ removes noise while leaving nearly all relevant events intact, creating a balance between noisy data and relevant features data. Further increasing the subsampling rate is not recommended, as it would considerably reduce the filter performance. The event filter with a subsampling rate of $s=4$ filters, on average, $60.76\%$ of all events (average value of all subjects and object scenarios). 

\begin{figure}[]
  \centering\includegraphics[width=1\textwidth]{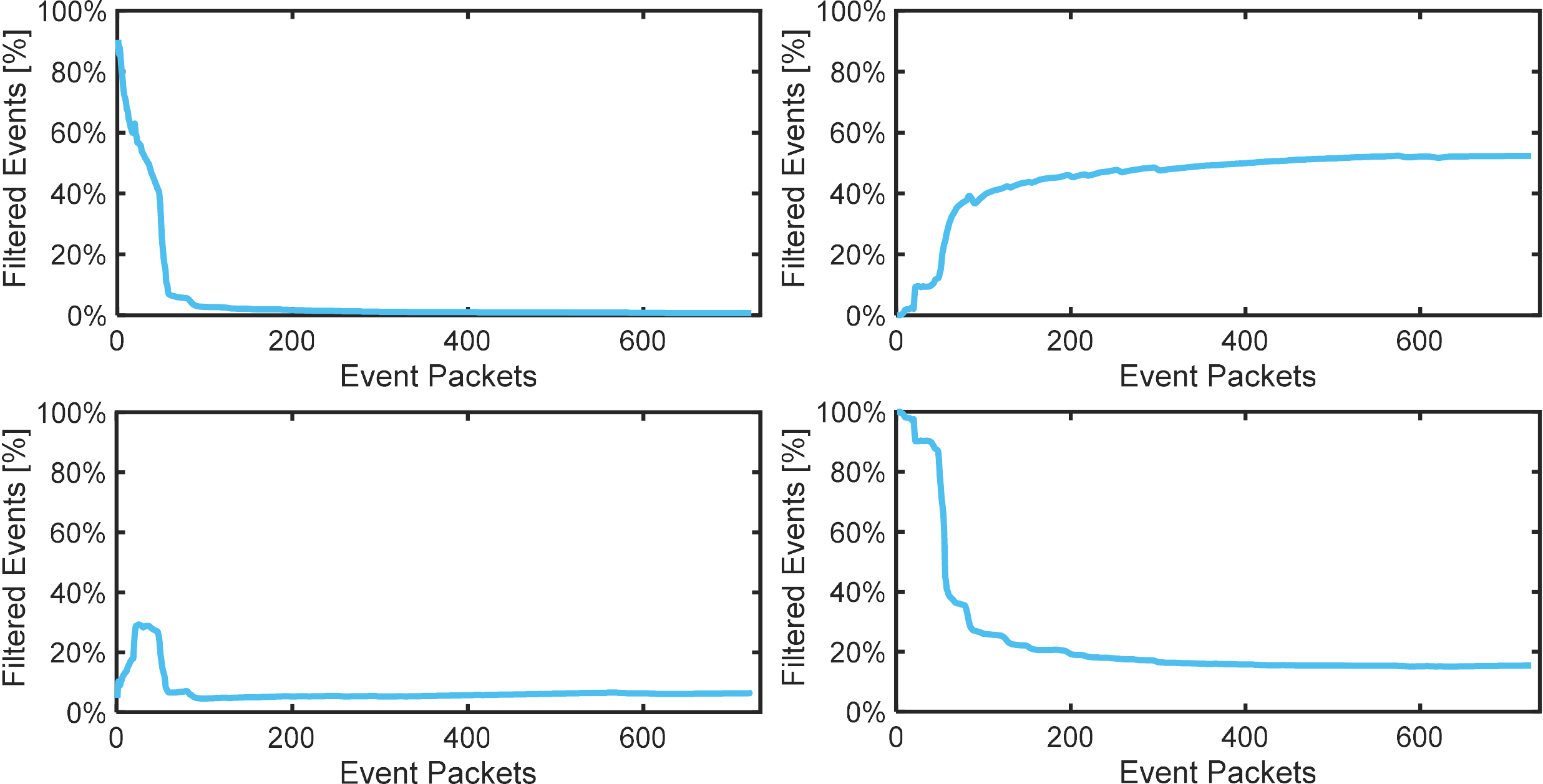}
\caption{Cumulative performance of individual filters (DMT22 dataset). First Event Filter (top left), Background Activity Filter (top right), Hot Pixel Filter (bottom left) and Refractory Filter (bottom right).}
\label{fig:12}       
\end{figure}

Considering an assembly sequence from the DMT22 dataset, the performance of each filter was evaluated by, at each event packet (collection of events grouped for processing purposes), comparing the number of total events received to the number of filtered events. By measuring these values cumulatively, at each new event packet, a graph is plotted showing the filter’s performance along the sequence, Fig.~\ref{fig:12}, showing its effectiveness over time. It is important to note that, due to measuring cumulatively, most graphs will show fewer fluctuations in value for the percentage of filtered events when near or at the end of the sequences. The results for each filter for the first sequence of the DMT22 dataset (first hard disk drive assembly by Subject 1) are presented in Fig.~\ref{fig:12}.

Both the First Event filter, Fig.~\ref{fig:12} (top left), and the Refractory filter, Fig.~\ref{fig:12} (bottom right), behave similarly to an exponential decay function, in the sense that they filter almost all initial events of the sequence, but quickly lessen their filtering performance. They are effective to filter most initial noise events, which, due to their abundance, create spatio-temporal neighbourhoods which would pass through the other filters. The Hot Pixel filter, Fig.~\ref{fig:12} (bottom left), also filters more events at the beginning of the action sequence, due to more noise which has similar behaviour to hot pixels. After that, the filter stabilizes the percentage of filtered events, indicative of actual hot pixels being consistently recognized and filtered. The Background Activity filter, Fig.~\ref{fig:12} (top right), filters the most events out of all the individual filters presented. It filters consistently along the whole action sequence, filtering an average of $~55\%$ of incoming events. The initial build-up is related to the filter requiring past events to verify spatio-temporal correlation and these events still need to be accumulated.

\begin{figure}[]
  \centering\includegraphics[width=1\textwidth]{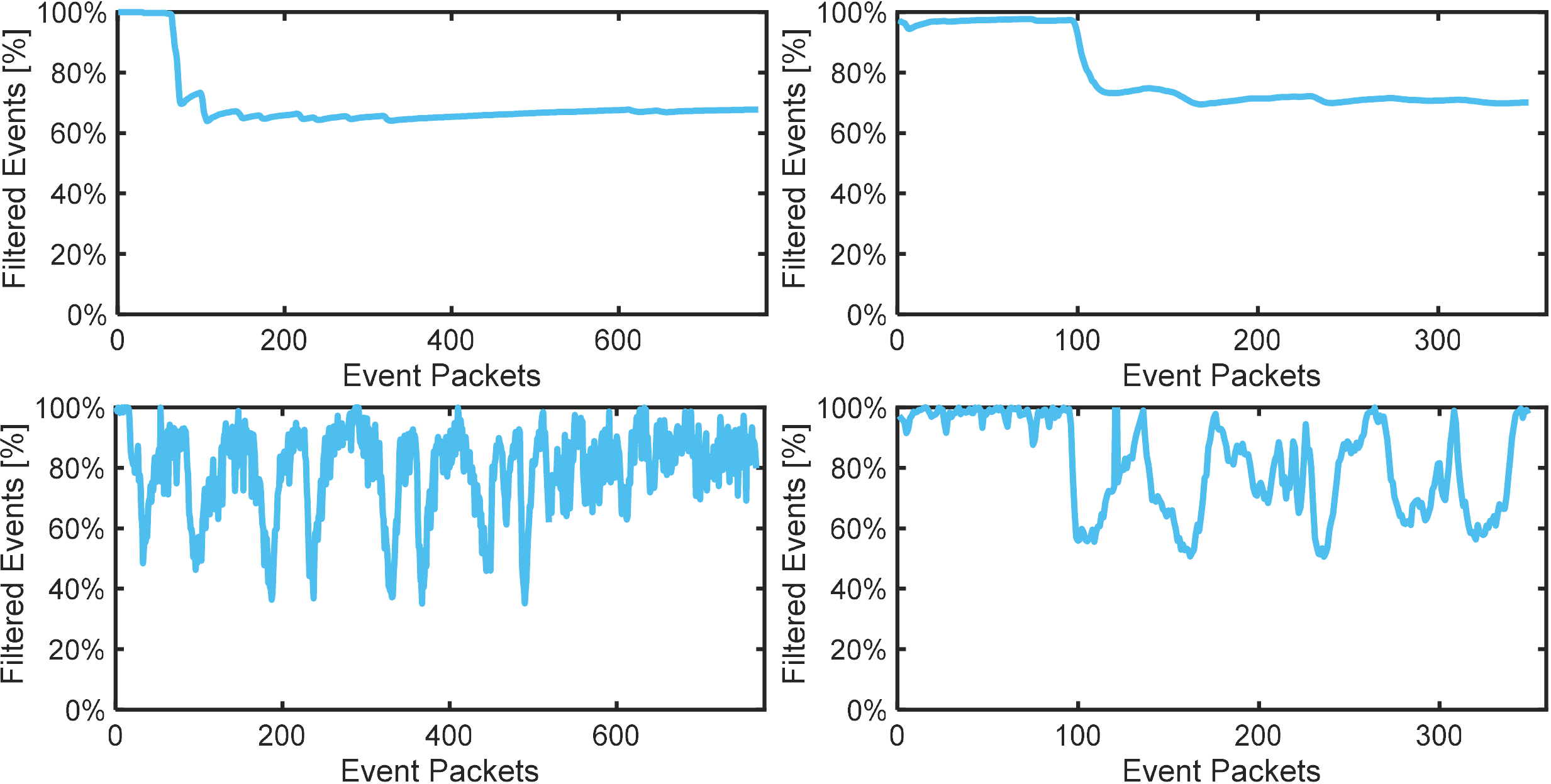}
\caption{Cumulative (top) and instantaneous (bottom) analysis of combined event filters on the DMT22 dataset (left) and on the EDHT21 dataset (right).}
\label{fig:13}       
\end{figure}

The progression of the cumulative percentage of filtered events along time, Fig.~\ref{fig:13} (top), mirrors the subject’s behaviour in the recorded sequence. At the beginning of the sequence, the subject is idle. During this period, all events that are output by the camera are noise and, ideally, the filters should act to filter them. At the moment the subject starts moving, an abrupt decrease in the percentage of filtered events can be observed. This decrease of about $~35\%$ is relative to the events created when the human subject moves the hand and object during the action. These events provide important information and should, in fact, not be filtered. Except for slight fluctuations, the percentage of filtered events during the actual sequence is mostly constant, due to the events created by noise being produced at a steady rate, and the hand moving at mostly the same speed and a consistent distance from the camera. The behaviour of combined filters during the first 3D sequence from the EDHT21 dataset, Fig.~\ref{fig:13} (top right), is very similar to that of the DMT22 dataset, Fig.~\ref{fig:13} (top left).

To better evaluate filters performance, a graph is plotted with instantaneous values of filter performance, Fig.~\ref{fig:13} (bottom). At each event packet, the total number of events received from the packet is compared to the number of events filtered from that same packet. The main difference between using the cumulative and instantaneous methods of plotting is that the latter shows a more descriptive representation of the filter performance during each assembly task. 

When the subject hand is moving, the percentage of filtered events decreases significantly. During the transition between different motion directions, the brief pause of movement of the subject can be identified in Fig.~\ref{fig:13} (bottom left) by the spikes at values of $100\%$ filtered events. In the EDHT21 dataset it is possible to identify specific changes in movement direction from the instantaneous behaviour of the event filter. In the sequence in Fig.~\ref{fig:13} (bottom right), the subject changes the direction of movement six times, which translates into a graph with six spikes at values of $100\%$ filtered events, although the exact location of each spike is not clearly defined. This behaviour, if consistently identified, can be used as a feature for action segmentation and, consequently, improve classification accuracy.

\subsection{Classification Results}
\label{sec:4.2}

\begin{figure}[]
  \centering\includegraphics[width=0.56\textwidth]{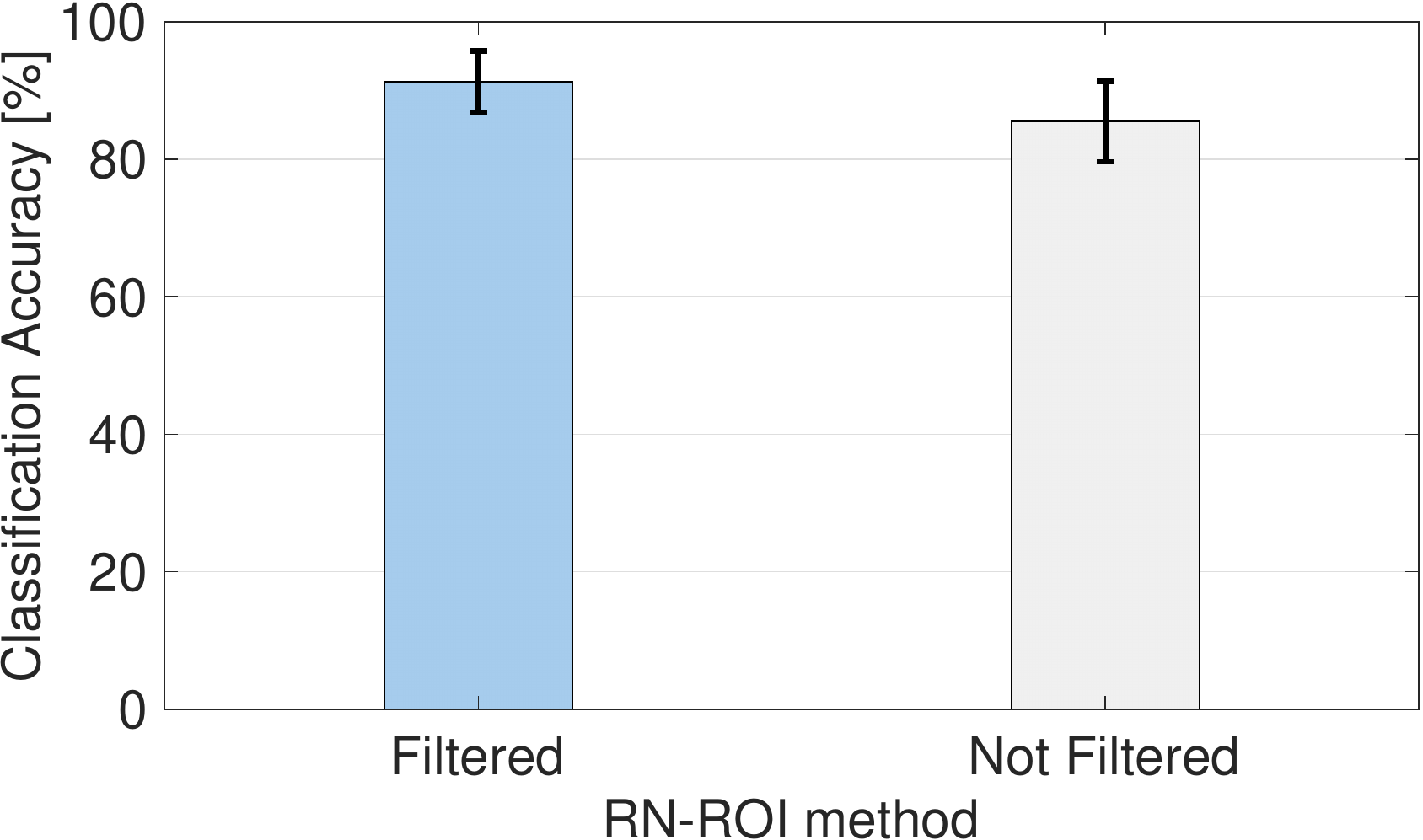}
\caption{Classification accuracy, with and without filters, applied on all the DMT22 event data, except left-sided object placement.}
\label{fig:15}       
\end{figure}

The filters performance was evaluated by measuring the general classification accuracy by feeding the RN-ROI with filtered and nonfiltered data. The LSTM network uses data from all five subjects in a division of $80\%$ training, corresponding to the data of 4 subjects, and $20\%$ testing, corresponding to the data of a single subject. In this particular test, only Subject 5’s right-hand object placement is used.
By isolating a specific subject’s data for testing, such data are uncorrelated to the training data and the network has no previous knowledge of this particular subject. The results obtained through this method are expected to be worse than those obtained through a random selection of data, but they are much more meaningful due to resembling a real-world scenario, where a new subject is introduced to the system. The subject that was chosen for testing is alternated between all possible combinations, as cross-validation, to test both the data and the network to their fullest. The general classification accuracies (average and the standard deviation) are in Fig.~\ref{fig:15}. They show that the use of event filters improves classification. The gain in classification accuracy is, approximately, $6$ percentage points, increasing from an average accuracy of $85.49\%$ to an average accuracy of $91.28\%$. The standard deviation error is reduced when using the filters. For all network runs, the filtered result accuracy always surpassed the non-filtered result.

A second set of tests was conducted to observe the classification accuracy in the context of different data selection scenarios: 
\begin{enumerate}
  \item Random sample selection featuring $85\%$ training data and $15\%$ testing data from the dataset. This is a common classification accuracy metric, where testing data are not available as training data, but the system is trained with data from Subject 5, which might bias the results. 
  \item Subject 5’s left-sided setup uses training data from all subjects, except Subject 5, and uses testing data from Subject 5’s left-sided setup (Subject 5’s left-hand object placement). The classifier has no previous knowledge of Subject 5.
  \item Subject 5’s right-sided setup uses training data from all subjects, except Subject 5, and uses testing data from Subject 5’s right-sided setup (Subject 5’s right-hand object placement). The classifier also has no previous knowledge of Subject 5.
\end{enumerate}

\begin{figure}[]
  \centering\includegraphics[width=0.85\textwidth]{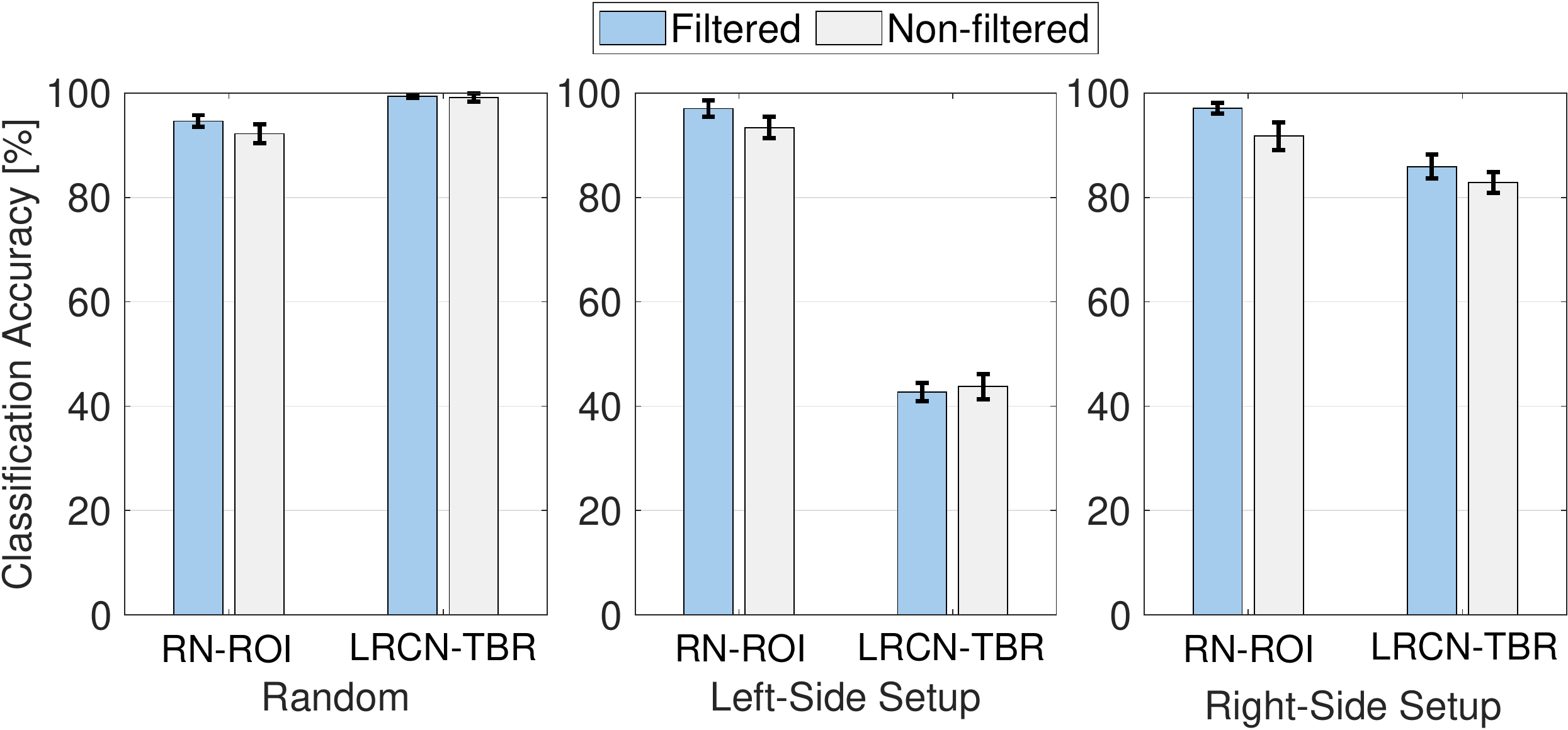}
\caption{Comparison of classification accuracy results from RN-ROI and LRCN-TBR methods, with and without filtered data. Three distinct scenarios were considered: randomly selected data (left), data from Subject 5's left-side setup (middle) and data from Subject 5's right-side setup (right).}
\label{fig:16}       
\end{figure}

For each of the scenarios above, the RN-ROI and LRCN-TBR classification methods were evaluated. The results in Fig.~\ref{fig:16} show that, for almost all cases, accuracy improves by using the event filters. Also, the variability of results, characterized through the standard deviation error, reduces when using filtered data. The overall best classification results are obtained for the filtered random selection scenario, with the LRCN-TBR’s $99,37\%$ accuracy outperforming the $94,63\%$ accuracy from RN-ROI. The classification accuracies for Subject 5’s recordings are very high when using the RN-ROI classifier, proving that the system is reliable to be used by right-handed and left-handed subjects, even after training the system without data from left-handed subjects. For the RN-ROI approach, the classification results of Subject 5’s left-sided and right-sided setup recordings are identical, with accuracies of $97,03\%$ and $97,08\%$, respectively. This suggests that, when using the mirror data augmentation technique, it is favourable to accommodate the left-handed subject with a left-sided configuration of objects, without loss of classification accuracy. For the scenarios of the left-side setup and right-side setup, the classification accuracy significantly degrades when using LRCN-TBR method, especially for the left-side setup. In such a context, it can be concluded that deep learning based LRCN-TBR method is more dependent on the training data from a specific subject than RN-ROI.

\begin{figure}[]
  \centering\includegraphics[width=0.5\textwidth]{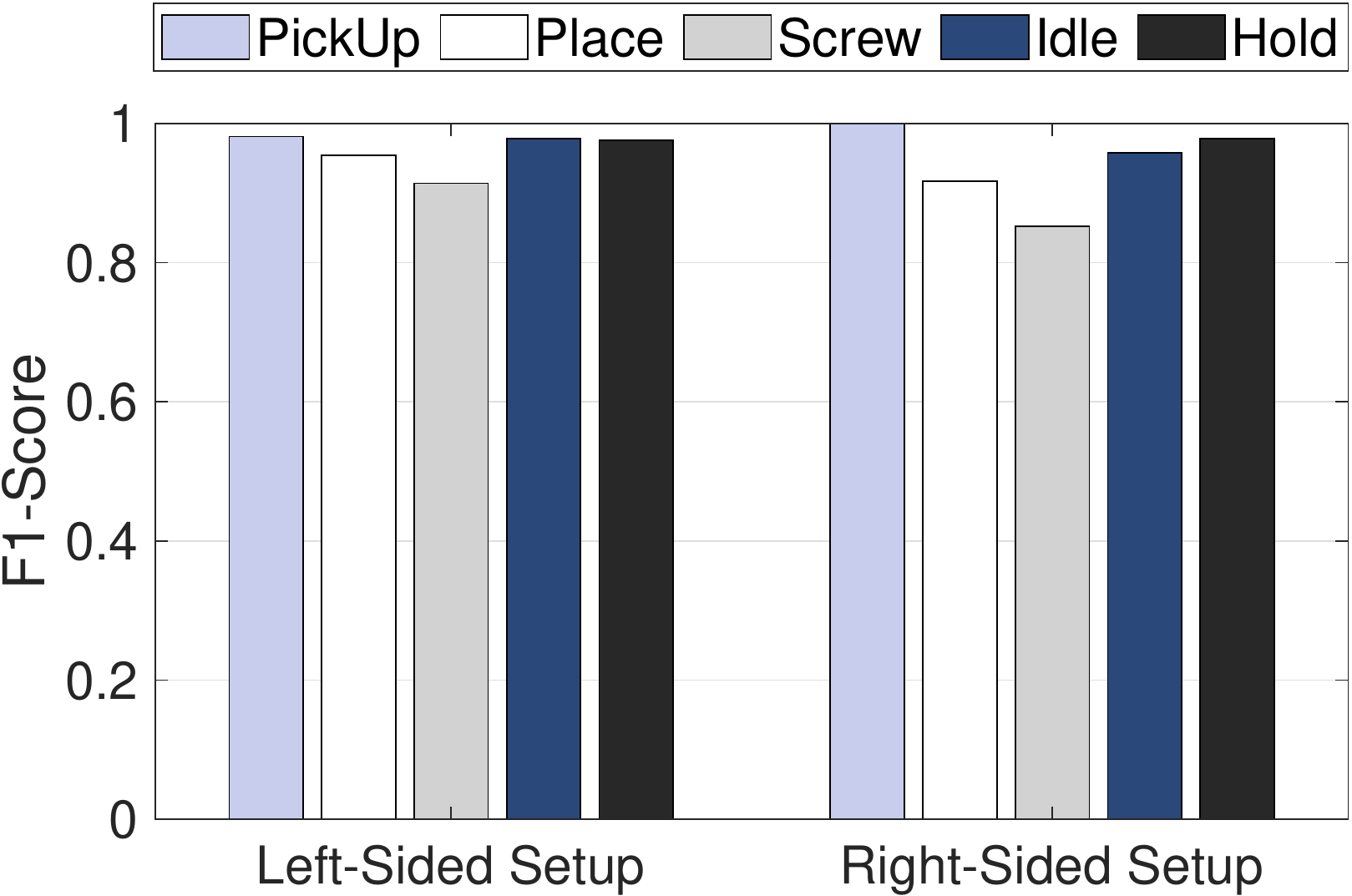}
\caption{F1-score per class from the DMT22 dataset using the RN-ROI method.}
\label{fig:17}       
\end{figure}

The F1-score is chosen as an auxiliary metric to evaluate classification performance due to the imbalance in the class distribution of the DMT22 dataset. The F1-score evaluates precision and recall, comparing the number of correct guesses (TP) against the number of other tasks which are misclassified as the intended class (FP) or against the number of intended tasks which are misclassified as another task (FN). The F1-score is calculated for each class from the DMT22 dataset, demonstrating similar accuracy when using RN-ROI, Fig.~\ref{fig:17}.

\begin{equation}
{F1}=\frac{2\times{precision}\times{recall}}{precision+recall} 
\end{equation}

\begin{equation}
{precision}=\frac{TP}{TP+FP}
\end{equation}

\begin{equation}
{recall}=\frac{TP}{TP+FN}
\end{equation}

\begin{figure}[]
  \centering\includegraphics[width=0.5\textwidth]{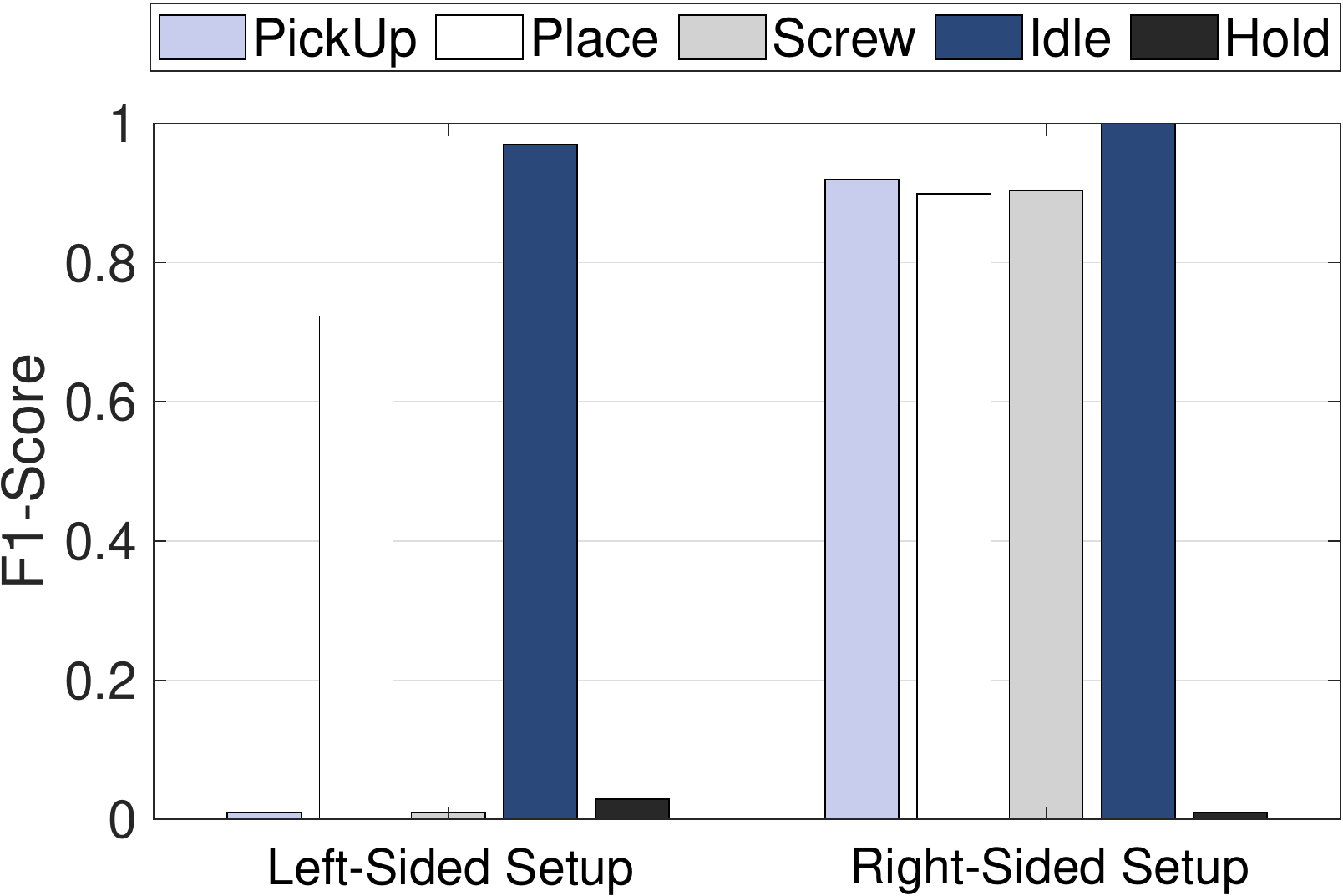}
\caption{F1-score per class from the DMT22 dataset using the LRCN-TBR method.}
\label{fig:18}       
\end{figure}

 The LRCN-TBR method falls short of classifying data from new subjects (which are not represented in the training dataset), especially in the case of the left-sided setup where classification accuracy reached at most $43,77\%$ using non-filtered data. One reason for this behaviour might be that CNN-based features can not characterize movement, while tracking-based features, such as ROI features do. However, this does not explain why the left-sided and right-sided setups have very distinct classification accuracies. To further analyse this incongruency, the F1-score was calculated, per class, to compare the two setups for the LRCN-TBR method, Fig.~\ref{fig:18}. The Idle class is the only class that is always correctly classified and both setups struggle to classify the Hold task. This makes sense, as the Idle class has the most distinct frames out of all the tasks (almost no events). Also, the Hold task is the most difficult task to classify as it has the least data for the network to learn and features a lot of similarities to the Place task. The Place task is well classified in both setups. The left-sided setup struggles most in the PickUp and Screw tasks. The misclassifications in these cases are not consistent, which indicates that the network does not have a good grasp on the features which characterize these tasks.

\begin{figure}[]
  \centering\includegraphics[width=1\textwidth]{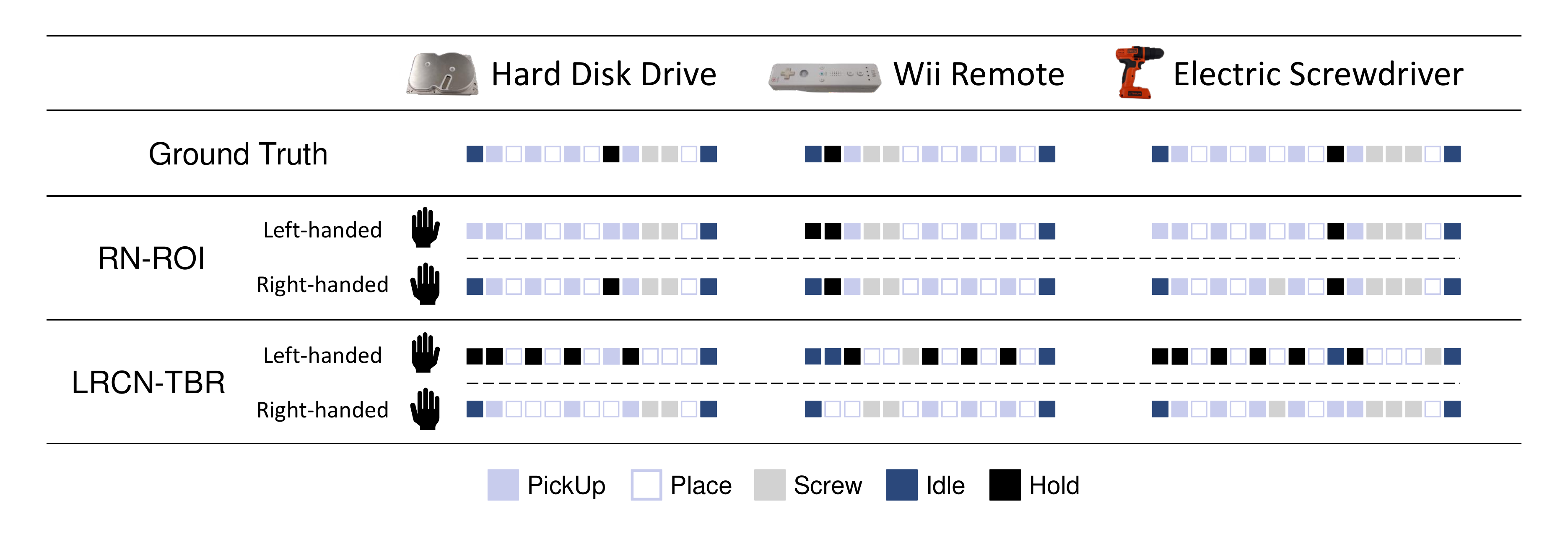}
\caption{Primitive assembly tasks classified in a continuous stream of data from the DMT22 dataset using the RN-ROI and LRCN-TBR methods. Results show two sequence lines for RN-ROI and LNCR-TBR, representing the classified task primitives when the subject testing the system is left-handed (top line) and right-handed (bottom line). Comparing the results with the ground truth primitive tasks, it is noted that the classification results obtained by the right-handed subject are better than the ones obtained by the left-handed subject, especially when using the LRCN-TBR method. This is likely because the DMT22 dataset is mostly composed by data collected from right-handed subjects.}
\label{fig:19}       
\end{figure}

Fig.~\ref{fig:19} shows task predictions for each object and method against the ground truth. Following previously discussed results, the Hold task is the most frequently misclassified.

\section{Conclusion and Future Work}
A novel methodology to classify manufacturing assembly primitives from event data was presented. Results demonstrated the effectiveness of the proposed deep learning and recurrent network classifiers, especially when event data are filtered. The combination of different filters promoted a dynamic response of the system, consistently targeting and removing noise events. On average, they filter out $65\%$ of the events from each recording. The classification accuracy, evaluated on the proposed DMT22 dataset, is about $91,31\%$ (using filters), $6$ percentage points higher than the classification accuracy obtained without using filters. In such a context, it can be concluded that the multiple filters play a key role in the classification accuracy when using event data as main input source. In addition, less data makes the classification faster and require less storage resources. It can also be concluded that in general the RN-ROI method presents better classification accuracy than the deep learning based LRCN-TBR method, especially when used by left-handed subjects who did not train the system. LRCN-TBR is more dependent on the training data from a specific subject than RN-ROI.

From a practical application perspective, the classified primitives can serve as input for a human-robot collaborative system that anticipates the co-worker’s needs (bringing parts and tools to the assembly workplace), learns from the co-worker’s demonstrations, and activates safety procedures according to the actual tasks being performed by the human co-worker. The proposed methodology has the added benefit of being object independent.

The proposed DMT22 dataset of manufacturing primitives should be extended and complemented with more data (novel assembly scenarios and new subjects). Movements that are not defined as actions, such as transition movement between tasks, should also be considered and labelled, aiming to obtain better classification and hand tracking results.

\section*{Acknowledgments}
This work was supported by FCT (Fundação para a Ciencia e a Tecnologia) [2021.06508.BD].

 \bibliographystyle{elsarticle-num} 
 \bibliography{article-refs}






\end{document}